\pdfoutput=1

\documentclass[11pt]{article}

\usepackage[]{ACL2023}

\usepackage{times}
\usepackage{latexsym}
\usepackage{enumitem}

\usepackage[T1]{fontenc}

\usepackage[utf8]{inputenc}

\usepackage{microtype}

\usepackage{inconsolata}
\usepackage{graphicx}
\usepackage{amsmath}
\usepackage{enumitem}

\usepackage{booktabs}

\definecolor{mypurple1}{RGB}{143, 94, 255}
\definecolor{myblue1}{RGB}{59, 76, 206}
\definecolor{myred1}{RGB}{190, 90, 75}
\definecolor{myyellow1}{RGB}{255, 192, 0}
\definecolor{mygreen1}{RGB}{160, 194, 128}

\title{Towards Robust Personalized Dialogue Generation via Order-Insensitive Representation Regularization}

\author{Liang Chen, Hongru Wang, Yang Deng, Wai-Chung Kwan, Zezhong Wang, Kam-Fai Wong \\
  The Chinese University of Hong Kong\\
  MoE Key Laboratory of High Confidence Software Technologies\\
  \tt \{lchen,hrwang,wckwan,zzwang,kfwong\}@se.cuhk.edu.hk \\
   \tt \{dengyang17dydy\}@gmail.com
}

\begin{document}

\maketitle
\begin{abstract}

Generating persona consistent dialogue response is important for developing an intelligent conversational agent.
Recent works typically fine-tune large-scale pre-trained models on this task by concatenating persona texts and dialogue history as a single input sequence to generate the target response. 
While simple and effective, our analysis shows that this popular practice is seriously affected by \textit{\textbf{Order Sensitivity}} where different input orders of persona sentences significantly impact the quality and consistency of generated response, resulting in severe performance fluctuations (i.e., 29.4\% on GPT2 and 83.2\% on BART).  
To mitigate the order sensitivity problem, we propose a model-agnostic framework, \textbf{OR}der \textbf{I}nsensitive \textbf{G}eneration (\textbf{ORIG}), which enables dialogue models to learn robust representation under different persona orders and improve the consistency of response generation. 
Experiments on Persona-Chat dataset justify the effectiveness and superiority of our method with two dominant pre-trained models (GPT2 and BART).\footnote{The code is available at \url{https://github.com/ChanLiang/ORIG}.}

\end{abstract}

\section{Introduction}

Developing a persona-consistent dialogue model has been one of the key issues and crucial problems in open-domain dialogue systems \citep{huang2020challenges}. 
\citet{zhang_personalizing_2018} define the problem of personalized dialogue generation, which aims to generate personalized responses based on textually described persona profiles. 
Many efforts have been made on developing dialogue models that generate responses consistent with the provided persona profile \cite{song_exploiting_2019,acl20-persona-gdr,song_generating_2020,acl20-persona-variational}.

\begin{figure}
    \centering
    \includegraphics[width=\linewidth]{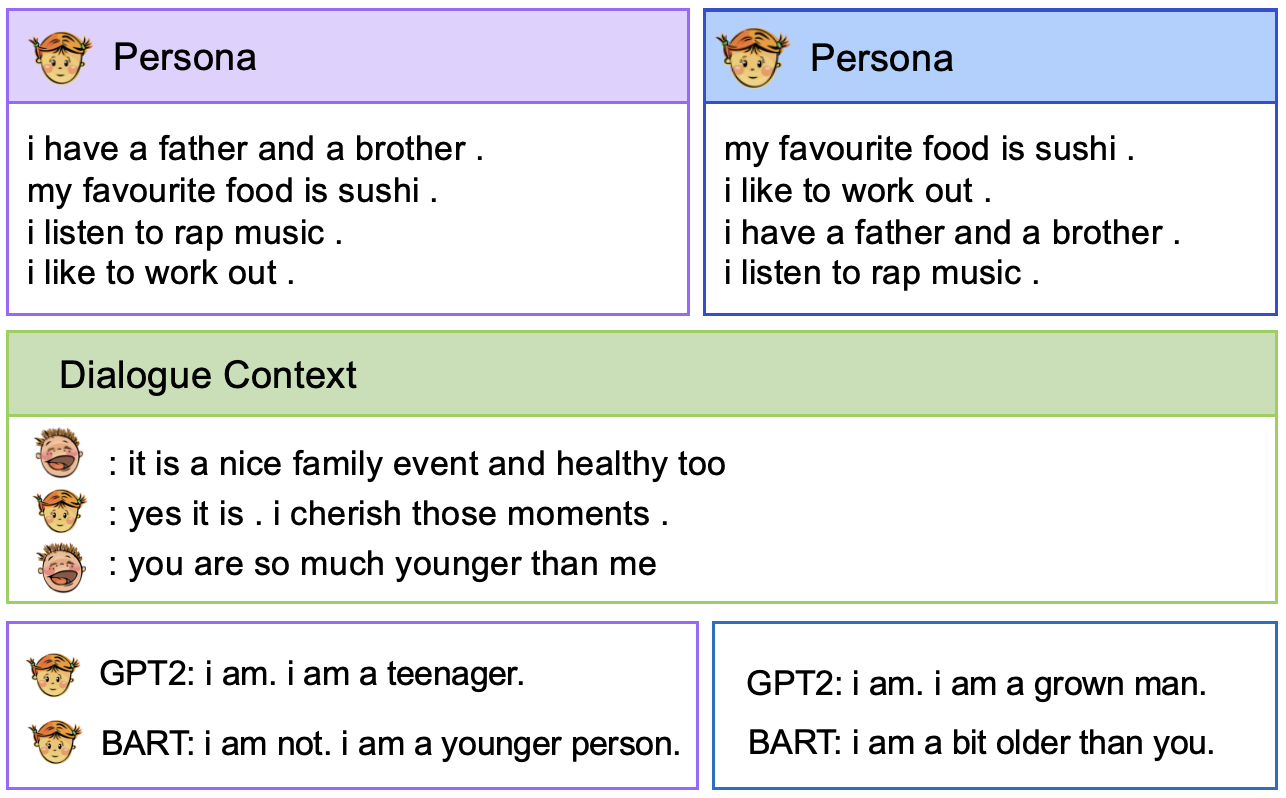}
    \caption{A dialog extract from Persona-Chat showing different orderings of the same persona can lead to different and even inconsistent responses.}
    \label{fig:intro-case}
\end{figure}

The recent development in transformer-based pre-trained models \cite{NIPS2017_3f5ee243,devlin2018bert,liu2019roberta,map-bert} has led to great successes in dialogue systems \citep{wolf_transfertransfo_2019,wu-etal-2020-tod,ham-etal-2020-end,kulhanek-etal-2021-augpt,cao-etal-2022-model,www22-use,tois23-crs,acl23-esc}.  
Inspired by these successes, previous works incorporate those pre-trained models in persona-based response generation by concatenating the dialogue history and persona as input to generate the response in an auto-regressive manner \cite{song-etal-2021-bob,liu_improving_2022}. However, a fine-tuned model can generate a high-quality and persona-consistent response in a certain ordering of personas, while varying this order may lead to a generic and even inconsistent response as illustrated by the example in Figure~\ref{fig:intro-case}. We empirically show that the worst ordering of persona can lead to a 29.4\% decline in BLEU score compared with the best ordering.

Ideally, a well-trained dialogue generation model should be able to generate a persona-consistent response regardless of the ordering of personas in the input.
We perform experiments and analyses to identify the cause of the ordering sensitivity. 
We find that the ordering of persona in the input leads to different representations of context and response. We also show that the model can attend to the appropriate persona and generate high-quality responses under some representations but not under others. This leads to instability in response generation.


Motivated by the above findings, we propose \textbf{OR}der \textbf{I}nsensitive \textbf{G}eneration (\textbf{ORIG}), which is a simple and effective framework that helps models learn more robust and better representations for different persona orders. More specifically, we formulate ORIG as a constrained optimization problem, which optimizes a persona response generation objective under the constraint: given different orderings of persona, the response representations of the model are the same. Then we optimize it through a stochastic optimization approach.

Experimental results on the Persona-Chat dataset show that ORIG significantly improves the robustness of pre-trained models (GPT2~\cite{radford2019language} and BART~\cite{lewis-etal-2020-bart}) under different orderings of input persona, as well as advances their generation performance.

In summary, our contributions are threefold: 
(1) We identify the order sensitivity problem in persona dialogue generation and conduct an empirical analysis to reveal its underlying reasons. 
(2) We propose a model-agnostic framework, ORIG, that helps different persona dialogue models learn robust representations while achieving better performance.
(3) We perform extensive experiments on the Persona-Chat dataset, showing that ORIG outperforms previous models and is more robust and less sensitive to different persona orderings.

\section{Related Work}

Maintaining a consistent persona is essential for building a human-like dialogue system, where most works regard persona as a set of sentences along with each dialog \citep{zhang_personalizing_2018,gu-etal-2019-dually,song_exploiting_2019,wu-etal-2021-transferable,cao-etal-2022-model,tois22-pqa}. \citet{song-etal-2021-bob} disentangled the task of persona-based dialogue generation into two sub-tasks: consistency understanding and dialogue generation while \citet{cao-etal-2022-model} aims to alleviate the problem of limited data by data manipulation methods. Despite satisfactory performance in previous work, the impacts of different orders of personas are still under-explored, resulting in unstable and inconsistent responses.

Our work is also related to work on order sensitivity in prompt-based few-shot learning \citep{zhao2021calibrate,lu-etal-2022-fantastically}. \citet{zhao2021calibrate} found that the different order of training examples in the prompt can cause accuracy to vary from near chance to state-of-the-art in the few-shot classification setting. 
Similarly, order sensitivity for In-context Learning also exists regardless of model size and the prompt format \citep{lu-etal-2022-fantastically}. 
Distinguishing from them, we focus on order sensitivity in the language generation task in finetuning setting, especially the impacts of persona orderings to generate persona-consistent responses.

\section{Order Sensitivity Problem and Analysis}\label{sec:analysis}

In this section, we first illustrate the seriousness of the order sensitivity problem by showing a huge performance fluctuation in persona dialogue models when fed the same personas in the best and worst orders. Then we analyse why their performance is volatile to different persona orderings.

To illustrate the problem, we finetune PLMs on the Persona-Chat by concatenating the persona and dialogue context together to predict the target response, including GPT2 and BART. After the training converges, we test them on two settings: 
(1) the best case: for each test sample, we feed the models all possible permutations of persona sentences and keep the maximum score for each sample as the final score; (2) the worst-case: perform the same process as (1), but take the minimum score. Table~\ref{tab:case_talbe} shows the results for two models. 
Surprisingly, we find the ordering of input persona has a big impact on the models' performance: GPT2’s worst case is 29.4\% lower than its best case, while BART’s is 83.2\% lower.

\begin{table}[t!]
\setlength{\tabcolsep}{2.0mm}
\small
\centering
\begin{tabular}{l|cccc}
\toprule
\textbf{Model} & BLEU-1 & BLEU-2 & ROUGE & CIDEr \\

\midrule
GPT2-best & 16.79 & 9.25 & 18.44 & 17.56  \\
GPT2-worst & 11.85 & 5.83 & 11.79 & 5.51 \\

\midrule
BART-best. & 28.17 & 18.29 & 31.07 & 46.53   \\
BART-worst. & 4.73 & 1.99 & 4.37 & 1.34   \\

\bottomrule
\end{tabular}
\caption{\label{tab:case_talbe} Performance gap between the best case and worst case when changing the ordering of input persona. }
\end{table}

Moreover, we find that the huge fluctuation in models' performance is closely related to the response representation changes caused by different orderings of input persona sentences. Concretely, we measure the similarity of the responses representation of the same test sample under different input orders of persona. We show their token-level similarity in the Table~\ref{tab:analysis_table} (persona and context are omitted for brevity), where the bidirectional KL function is employed as the distance function. 
Ideally, models should have the consistent response representation: KL distance between the same response should be zero.
However, their distances are significantly higher than zero.
It reveals that the models behave more likely a left-to-right language model whose representation is prone to the different orderings of the previous input (e.g. persona sentences). That is highly undesirable for a robust personalized dialogue model.
Thus, regularization of representation for the response tokens is necessary to help personalized dialogue models capture order-invariant representation.

\begin{table}[]
\footnotesize
\begin{tabular}{c|p{170pt}}
\hline
\multicolumn{1}{c|}{BART}        & great(0.185) and(0.105) how(0.312) was(0.289) your(0.124) day(0.304) ? \\
\hline

\end{tabular}
\caption{The token-level representation of the same response can be very different when the ordering of input persona changes. The value denotes the KL distance of the same tokens representation returned by the models fed with two different orderings of persona.}  
\label{tab:analysis_table}
\end{table}

\section{Method}

\begin{figure}[t]
\centering
\includegraphics[trim={3.5cm 7cm -13cm 5cm}, clip, scale=0.5, width=1.0\textwidth]{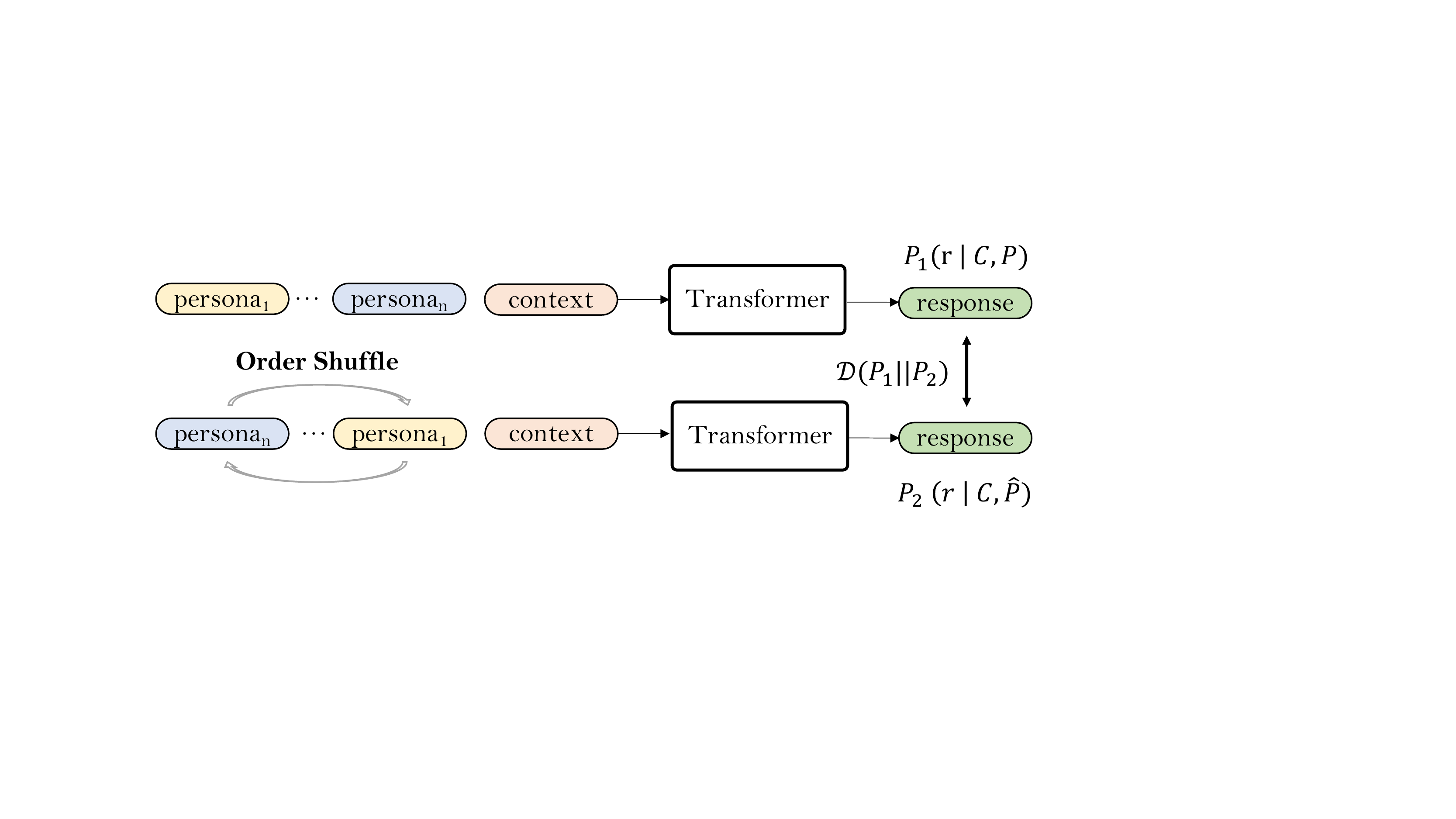}
\captionsetup{font=10pt}
\caption{Our proposed framework ORIG}
\label{model}
\end{figure}

We introduce the proposed framework, named \textbf{ORIG}: \textbf{OR}der \textbf{I}nsensitive \textbf{G}eneration (\textbf{ORIG}). As shown in  Figure~\ref{model}, we transform the persona order-sensitivity problem as a constrained optimization problem that optimises a persona dialogue model under the uncertainty of the input persona order.

\subsection{Problem Formulation}

Given the dialogue context $C=\{u_1,\dots,u_m\}$ and a set of persona descriptions $P=\{p_1,\dots,p_n\}$, the goal is to generate a personalized response $r$. Formally, the generation problem can be formulated as the following chain rule:
\begin{equation}
P(r | C, P ; \theta)= \prod\nolimits_{t=1}^{T} P\left( r_t | r_{1:t-1}, C, P ; \theta\right)
\label{dialog_persona}
\end{equation}
where $\theta$ is the parameters of the dialogue model.

\subsection{ORIG Framework}
\label{method1}

According to the analysis in Section~\ref{sec:analysis}, the observation reveals that varying the order of input personas leads to different representations of the dialogue response, thus resulting in fluctuations in performance.

To learn more robust and consistent representations, we propose the ORIG framework that complements the response generation process with a constraint: given the different orderings of a persona, the model's response representations need to be the same.

Then the order-insensitive personalized dialogue generation problem is modelled as the following constrained optimization problem

\begin{align}
    \min_\theta [-\log P(r|C, P; \theta)] \label{eq:2}\\
    \textit{s.t.} \quad \mathcal{D}[P(r|C, P;\theta), P(r|C, \hat{P};\theta))]=0 \label{eq:3} \\
    (P, C, r)\sim D \label{eq:5} \\
    \hat{P}\sim\text{Shuffle}(P) \label{eq:6}
\end{align}

where $P(r|C, P; \theta)$ are the model's predictions over the dialogue response, D denotes the dialogue corpus, and the function $\mathcal{D}$ is KL divergence to measure the difference between two distributions, and the Shuffle operator samples each persona ordering uniformly from the full permutation of $P$.

\subsection{Optimization}

\label{method4}

As for optimization, we first apply the Lagrange multipliers strategy to convert the constrained problem into an unconstrained problem

\begin{equation} 
\begin{aligned}
    \mathcal{L}_\theta = &  -\log P(r|C, P; \theta) \\
     &\left.  + \gamma \cdot \mathcal{D}[P(r|C, P; \theta), P(r|C, \hat{P}; \theta)] \right.
\label{eq:gpt2}
\end{aligned}
\end{equation}
where $\gamma$ is the multiplier corresponding to the equality constraints (\ref{eq:3}). Then we can update
the parameters $\theta$ of dialogue models by stochastic gradient descent.

\section{Experiments}

\subsection{Experimental Setups}
\textbf{Datasets}~ We evaluate the models on the Persona-Chat dataset \citep{zhang_personalizing_2018}, where each dialogue session has at least 6 turns of interactions. And each interaction is conditioned on a persona that is described with 5 profile sentences.

\begin{table*}[t!]
\setlength{\tabcolsep}{1.5mm}
\small
\centering
\begin{tabular}{l|cccccc|ccc}
\toprule
 & \multicolumn{6}{c}{Automatic Evaluations} & \multicolumn{3}{c}{Human Evaluations} \\
 
\midrule
\textbf{Model} & BLEU-1 & BLEU-2 & ROUGE & Entropy & CIDEr & \textbf{C}-score & Flu. & Con. Coh. & Per. Cons. \\

\midrule
\textsc{GPT2} & 13.95 & 7.22 & 14.82 & 6.53 & 10.10 & 0.718 & 1.531 & 1.281 & 1.719 \\

\textsc{GPT2-ORIG} & \bf 14.61 & \bf 7.43 & \bf 14.94 & \bf 6.54 & \bf 10.60 & \bf 0.733 & \bf 1.726 & \bf 1.512 & \bf 1.719 \\

\midrule
\textsc{BART} & 14.19 & 7.61 & 15.05 & \bf 6.67 & 11.07 & 0.443 & 1.906 & 1.312 & 1.438 \\

\textsc{BART-ORIG} & \bf 14.64 & \bf 7.90 & \bf 15.20 & 6.41 & \bf 13.27 & \bf 0.446 & \bf 1.938 & \bf 1.332 & \bf 1.457 \\

\bottomrule
\end{tabular}

\caption{\label{tab:normal_order}Automatic and human evaluation results of applying ORIG on two base models in the original test set without any modifications on input persona orders. }

\end{table*}

\noindent\textbf{Baselines}~ To verify the generality of our framework across different architectures, we perform experiments on the two most popular pre-trained architectures: Transformer encoder-decoder (BART) and Transformer decoder (GPT2).

\noindent\textbf{Implementation Details}~ We choose GPT2 base (117M) and BART base (139M) as the base models and compare the base models finetuned with classical max likelihood estimation (MLE) and our proposed ORIG. Our implementation was based on HuggingFace's Transformers library~\cite{wolf-etal-2020-transformers}. During training, the learning rate is set as $2 \times 10^{-5}$, and the batch size for GPT2 and BART is set as 64 and 32, respectively. We trained both models for 10 epochs with Adam~\cite{kingma2014adam} optimizer until they converged. During decoding, We employ a top-p (p=0.9)~\cite{holtzman2020curious} plus top-k (k=50) sampling strategy, which is used to avoid sampling from the unreliable tail of the distribution (only consider a subset of vocabulary composed of k words with the highest probability or some most probable words whose sum of probabilities equals p at each decoding step). The random seed for all experiments is set to 42. 

\noindent\textbf{Evaluation Metrics}~ We perform both automatic and human evaluations. 
(1) Automatic metrics: We adopt BLEU~\cite{papineni-etal-2002-bleu}, ROUGE~\cite{lin-2004-rouge}, Entropy\cite{zhang_entropy} and CIDEr~\cite{vedantam2015cider} for lexical-based measurement.
 Following previous work, we also adopt the C-score~\cite{cscore} to indicate the consistency between the generated response and the input personas. C-score is calculated by the entailment score of a RoBERTa model finetuned on the DialogueNLI dataset.
(2) Human evaluation: We randomly sampled 200 samples from the test set and ask 3 crowdworkers to rate the generated responses in the following three aspects:  response fluency, context coherence and persona consistency. The scores $\{0, 1, 2\}$ indicate unacceptable, acceptable and excellent, respectively. The degree of agreement during human evaluation is measured by Fleiss' kappa \cite{fleiss1971measuring}.

 \begin{figure}
    \centering

    \includegraphics[width=0.40\textwidth]{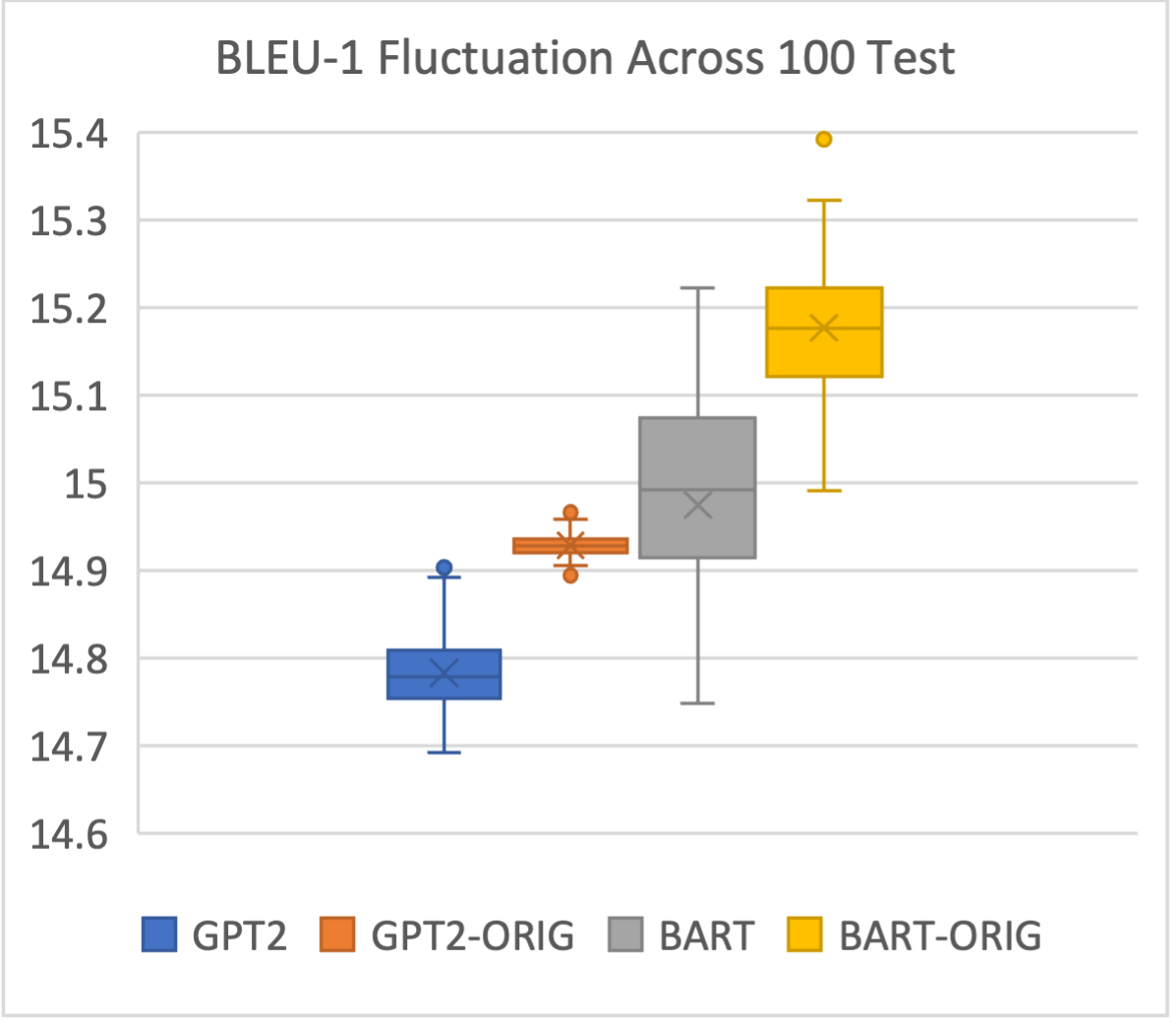}

    \caption{Our ORIG improves the mean performance while reducing the variance of both models. Statistics about running each model 100 times on the test set and randomly shuffling the order of the input persona sentences in each run.}
    \label{fig:gpt2-box}
\end{figure}

\begin{table}[h!]
\setlength{\tabcolsep}{2.5mm}
\small
\centering
\begin{tabular}{l|cc|cc}
\toprule
 \textbf{Model} & mean & variance & best & worst \\
 
\midrule
\textsc{GPT2} & 14.78 & 0.00193 & \bf 16.79 & 11.85 \\
\textsc{GPT2-ORIG} & \bf 14.93 & \bf 0.00016 & 14.95 & \bf 14.25 \\
\midrule
\textsc{BART} & 15.01 & 0.01123 & \bf 28.17 & 4.73 \\
\textsc{BART-ORIG} & \bf 15.18 & \bf 0.00532 & 26.44 & \bf 5.80 \\
\bottomrule
\end{tabular}
\caption{\label{tab:statistics_} Statistical results of BLEU-1. The mean and variance are obtained by running each model 100 times on the test set and randomly shuffling the order of the input persona sentences in each run. The best case and worst case are obtained by feeding models the best and worst orderings of personas for every test sample.
}
\end{table}

\subsection{Experimental Results}

 \textbf{Improves performance in the original test set}~ Table~\ref{tab:normal_order} shows different models' performance in the original test set without any modifications (for ORIG, "Shuffle" is used during training but is optional during testing. The Table 3 caption signifies the absence of "Shuffle" during testing. This is to evaluate if ORIG performs well in the normal setting). From automatic metrics, we can see base models trained with our ORIG framework outperform the baselines. It justifies that our framework can be applied to different models to improve their performance. 
From human evaluation results, 
models with ORIG are superior to  others on almost all metircs, especially on GPT2. This is consistent with the results of automatic metrics. The average kappa value of the annotation is 0.632, indicating good agreement during human evaluation.

 \noindent\textbf{Reduces variance and improves mean and worst-case performance}~ Figure~\ref{fig:gpt2-box} shows that aside from reducing the variance, ORIG also improves mean and worst-case performance (detailed results in Table~\ref{tab:statistics_}) across two models consistently, especially in GPT2 (the worst case performance is very close to the best case). We reduce the variance on GPT2 and BART by 91.6\% and 51.8\%, respectively. Meanwhile, we improve worst-case performance by 20.3\% and 22.6\% on GPT2 and BART respectively. The only drop is the best case. This is because our distance function $\mathcal{D}$ is unidirectional, which pulls in the two representations in Equation~\ref{eq:3} indiscriminately, causing the best case to go down and the worst to go up. We leave more complicated and directional distance constraints for future studies.

\section{Conclusion}
We show that the current practice of applying pre-trained models to the personalized dialogue generation task is volatile across different input orders of personas. 
Through the analysis, we find that the problem arises from the representation changes induced by the input changes. Motivated by these, we propose our ORIG, a model-agnostic framework for finetuning the persona dialogue model such that it obtains a persona order-invariant representation. Experiments on two dominant pre-trained dialogue models show that our framework improves performance and reduces order volatility.

\section*{Limitations}
In this section, we discuss the limitations of this work. First, on the problems side, it's non-trivial to consider the order of all kinds of grounding knowledge, but we have only explored Persona-Chat. We hope to apply our method to more grounded generation tasks such as knowledge-grounded and document-grounded dialogue in the future. Second, on the methods side, our framework is training-based, but we hope more lightweight techniques could be developed to improve the model's robustness even without training the model.


\section*{Acknowledgements}
We would like to thank Professor Helen Meng and Xixin Wu for their helpful discussion and feedback on the course SEEM5640. We also thank anonymous reviewers for their constructive comments. Thanks to Dr Honshan HO for his support. This research work is partially supported by CUHK under Project No. 3230366.

\bibliography{acl2023}

\begin{thebibliography}{36}
\expandafter\ifx\csname natexlab\endcsname\relax\def\natexlab#1{#1}\fi

\bibitem[{Cao et~al.(2022)Cao, Bi, Fang, Shi, and Tao}]{cao-etal-2022-model}
Yu~Cao, Wei Bi, Meng Fang, Shuming Shi, and Dacheng Tao. 2022.
\newblock \href {https://doi.org/10.18653/v1/2022.acl-long.550} {A
  model-agnostic data manipulation method for persona-based dialogue
  generation}.
\newblock In \emph{Proceedings of the 60th Annual Meeting of the Association
  for Computational Linguistics (Volume 1: Long Papers)}, pages 7984--8002,
  Dublin, Ireland. Association for Computational Linguistics.

\bibitem[{Chen(2020)}]{map-bert}
Liang Chen. 2020.
\newblock \href {http://arxiv.org/abs/2008.05333} {Variance-reduced language
  pretraining via a mask proposal network}.
\newblock \emph{CoRR}, abs/2008.05333.

\bibitem[{Deng et~al.(2022{\natexlab{a}})Deng, Li, Zhang, Ding, and
  Lam}]{tois22-pqa}
Yang Deng, Yaliang Li, Wenxuan Zhang, Bolin Ding, and Wai Lam.
  2022{\natexlab{a}}.
\newblock \href {https://doi.org/10.1145/3507782} {Toward personalized answer
  generation in e-commerce via multi-perspective preference modeling}.
\newblock \emph{{ACM} Trans. Inf. Syst.}, 40(4):87:1--87:28.

\bibitem[{Deng et~al.(2022{\natexlab{b}})Deng, Zhang, Lam, Cheng, and
  Meng}]{www22-use}
Yang Deng, Wenxuan Zhang, Wai Lam, Hong Cheng, and Helen Meng.
  2022{\natexlab{b}}.
\newblock \href {https://doi.org/10.1145/3485447.3512020} {User satisfaction
  estimation with sequential dialogue act modeling in goal-oriented
  conversational systems}.
\newblock In \emph{{WWW} '22: The {ACM} Web Conference 2022}, pages 2998--3008.

\bibitem[{Deng et~al.(2022{\natexlab{c}})Deng, Zhang, Xu, Lei, Chua, and
  Lam}]{tois23-crs}
Yang Deng, Wenxuan Zhang, Weiwen Xu, Wenqiang Lei, Tat{-}Seng Chua, and Wai
  Lam. 2022{\natexlab{c}}.
\newblock \href {https://doi.org/10.48550/arXiv.2204.06923} {A unified
  multi-task learning framework for multi-goal conversational recommender
  systems}.
\newblock \emph{CoRR}, abs/2204.06923.

\bibitem[{Deng et~al.(2023)Deng, Zhang, Yuan, and Lam}]{acl23-esc}
Yang Deng, Wenxuan Zhang, Yifei Yuan, and Wai Lam. 2023.
\newblock Knowledge-enhanced mixed-initiative dialogue system for emotional
  support conversations.
\newblock \emph{arXiv preprint arXiv:2305.10172}.

\bibitem[{Devlin et~al.(2018)Devlin, Chang, Lee, and
  Toutanova}]{devlin2018bert}
Jacob Devlin, Ming-Wei Chang, Kenton Lee, and Kristina Toutanova. 2018.
\newblock Bert: Pre-training of deep bidirectional transformers for language
  understanding.
\newblock \emph{arXiv preprint arXiv:1810.04805}.

\bibitem[{Fleiss(1971)}]{fleiss1971measuring}
Joseph~L Fleiss. 1971.
\newblock Measuring nominal scale agreement among many raters.
\newblock \emph{Psychological bulletin}, 76(5):378.

\bibitem[{Gu et~al.(2019)Gu, Ling, Zhu, and Liu}]{gu-etal-2019-dually}
Jia-Chen Gu, Zhen-Hua Ling, Xiaodan Zhu, and Quan Liu. 2019.
\newblock \href {https://doi.org/10.18653/v1/D19-1193} {Dually interactive
  matching network for personalized response selection in retrieval-based
  chatbots}.
\newblock In \emph{Proceedings of the 2019 Conference on Empirical Methods in
  Natural Language Processing and the 9th International Joint Conference on
  Natural Language Processing (EMNLP-IJCNLP)}, pages 1845--1854, Hong Kong,
  China. Association for Computational Linguistics.

\bibitem[{Ham et~al.(2020)Ham, Lee, Jang, and Kim}]{ham-etal-2020-end}
Donghoon Ham, Jeong-Gwan Lee, Youngsoo Jang, and Kee-Eung Kim. 2020.
\newblock \href {https://doi.org/10.18653/v1/2020.acl-main.54} {End-to-end
  neural pipeline for goal-oriented dialogue systems using {GPT}-2}.
\newblock In \emph{Proceedings of the 58th Annual Meeting of the Association
  for Computational Linguistics}, pages 583--592, Online. Association for
  Computational Linguistics.

\bibitem[{Holtzman et~al.(2020)Holtzman, Buys, Du, Forbes, and
  Choi}]{holtzman2020curious}
Ari Holtzman, Jan Buys, Li~Du, Maxwell Forbes, and Yejin Choi. 2020.
\newblock \href {http://arxiv.org/abs/1904.09751} {The curious case of neural
  text degeneration}.

\bibitem[{Huang et~al.(2020)Huang, Zhu, and Gao}]{huang2020challenges}
Minlie Huang, Xiaoyan Zhu, and Jianfeng Gao. 2020.
\newblock Challenges in building intelligent open-domain dialog systems.
\newblock \emph{ACM Transactions on Information Systems (TOIS)}, 38(3):1--32.

\bibitem[{Kingma and Ba(2015)}]{kingma2014adam}
Diederik~P Kingma and Jimmy Ba. 2015.
\newblock Adam: A method for stochastic optimization.
\newblock In \emph{ICLR}.

\bibitem[{Kulh{\'a}nek et~al.(2021)Kulh{\'a}nek, Hude{\v{c}}ek, Nekvinda, and
  Du{\v{s}}ek}]{kulhanek-etal-2021-augpt}
Jon{\'a}{\v{s}} Kulh{\'a}nek, Vojt{\v{e}}ch Hude{\v{c}}ek, Tom{\'a}{\v{s}}
  Nekvinda, and Ond{\v{r}}ej Du{\v{s}}ek. 2021.
\newblock \href {https://doi.org/10.18653/v1/2021.nlp4convai-1.19} {{AuGPT}:
  Auxiliary tasks and data augmentation for end-to-end dialogue with
  pre-trained language models}.
\newblock In \emph{Proceedings of the 3rd Workshop on Natural Language
  Processing for Conversational AI}, pages 198--210, Online. Association for
  Computational Linguistics.

\bibitem[{Lewis et~al.(2020)Lewis, Liu, Goyal, Ghazvininejad, Mohamed, Levy,
  Stoyanov, and Zettlemoyer}]{lewis-etal-2020-bart}
Mike Lewis, Yinhan Liu, Naman Goyal, Marjan Ghazvininejad, Abdelrahman Mohamed,
  Omer Levy, Veselin Stoyanov, and Luke Zettlemoyer. 2020.
\newblock \href {https://doi.org/10.18653/v1/2020.acl-main.703} {{BART}:
  Denoising sequence-to-sequence pre-training for natural language generation,
  translation, and comprehension}.
\newblock In \emph{Proceedings of the 58th Annual Meeting of the Association
  for Computational Linguistics}, pages 7871--7880, Online. Association for
  Computational Linguistics.

\bibitem[{Lin(2004)}]{lin-2004-rouge}
Chin-Yew Lin. 2004.
\newblock \href {https://aclanthology.org/W04-1013} {{ROUGE}: A package for
  automatic evaluation of summaries}.
\newblock In \emph{Text Summarization Branches Out}, pages 74--81, Barcelona,
  Spain. Association for Computational Linguistics.

\bibitem[{Liu et~al.(2022)Liu, Wei, Liu, Mao, Fang, and
  Chen}]{liu_improving_2022}
Yifan Liu, Wei Wei, Jiayi Liu, Xianling Mao, Rui Fang, and Dangyang Chen. 2022.
\newblock \href {https://doi.org/10.1145/3511808.3557359} {Improving
  {{Personality Consistency}} in {{Conversation}} by {{Persona Extending}}}.
\newblock In \emph{Proceedings of the 31st {{ACM International Conference}} on
  {{Information}} \& {{Knowledge Management}}}, pages 1350--1359.

\bibitem[{Liu et~al.(2019)Liu, Ott, Goyal, Du, Joshi, Chen, Levy, Lewis,
  Zettlemoyer, and Stoyanov}]{liu2019roberta}
Yinhan Liu, Myle Ott, Naman Goyal, Jingfei Du, Mandar Joshi, Danqi Chen, Omer
  Levy, Mike Lewis, Luke Zettlemoyer, and Veselin Stoyanov. 2019.
\newblock Roberta: A robustly optimized bert pretraining approach.
\newblock \emph{arXiv preprint arXiv:1907.11692}.

\bibitem[{Lu et~al.(2022)Lu, Bartolo, Moore, Riedel, and
  Stenetorp}]{lu-etal-2022-fantastically}
Yao Lu, Max Bartolo, Alastair Moore, Sebastian Riedel, and Pontus Stenetorp.
  2022.
\newblock \href {https://doi.org/10.18653/v1/2022.acl-long.556} {Fantastically
  ordered prompts and where to find them: Overcoming few-shot prompt order
  sensitivity}.
\newblock In \emph{Proceedings of the 60th Annual Meeting of the Association
  for Computational Linguistics (Volume 1: Long Papers)}, pages 8086--8098,
  Dublin, Ireland. Association for Computational Linguistics.

\bibitem[{Madotto et~al.(2019)Madotto, Lin, Wu, and Fung}]{cscore}
Andrea Madotto, Zhaojiang Lin, Chien-Sheng Wu, and Pascale Fung. 2019.
\newblock \href {https://aclanthology.org/P19-1542/?ref=https://githubhelp.com}
  {Personalizing dialogue agents via meta-learning}.
\newblock In \emph{Proceedings of ACL 2019}, pages 5454--5459.

\bibitem[{Papineni et~al.(2002)Papineni, Roukos, Ward, and
  Zhu}]{papineni-etal-2002-bleu}
Kishore Papineni, Salim Roukos, Todd Ward, and Wei-Jing Zhu. 2002.
\newblock \href {https://doi.org/10.3115/1073083.1073135} {{B}leu: a method for
  automatic evaluation of machine translation}.
\newblock In \emph{Proceedings of the 40th Annual Meeting of the Association
  for Computational Linguistics}, pages 311--318, Philadelphia, Pennsylvania,
  USA. Association for Computational Linguistics.

\bibitem[{Radford et~al.(2019)Radford, Wu, Child, Luan, Amodei, Sutskever
  et~al.}]{radford2019language}
Alec Radford, Jeffrey Wu, Rewon Child, David Luan, Dario Amodei, Ilya
  Sutskever, et~al. 2019.
\newblock Language models are unsupervised multitask learners.
\newblock \emph{OpenAI blog}, 1(8):9.

\bibitem[{Song et~al.(2021)Song, Wang, Zhang, Zhang, and
  Liu}]{song-etal-2021-bob}
Haoyu Song, Yan Wang, Kaiyan Zhang, Wei-Nan Zhang, and Ting Liu. 2021.
\newblock \href {https://doi.org/10.18653/v1/2021.acl-long.14} {{B}o{B}: {BERT}
  over {BERT} for training persona-based dialogue models from limited
  personalized data}.
\newblock In \emph{Proceedings of the 59th Annual Meeting of the Association
  for Computational Linguistics and the 11th International Joint Conference on
  Natural Language Processing (Volume 1: Long Papers)}, pages 167--177, Online.
  Association for Computational Linguistics.

\bibitem[{Song et~al.(2020{\natexlab{a}})Song, Wang, Zhang, Liu, and
  Liu}]{acl20-persona-gdr}
Haoyu Song, Yan Wang, Weinan Zhang, Xiaojiang Liu, and Ting Liu.
  2020{\natexlab{a}}.
\newblock \href {https://doi.org/10.18653/v1/2020.acl-main.516} {Generate,
  delete and rewrite: {A} three-stage framework for improving persona
  consistency of dialogue generation}.
\newblock In \emph{Proceedings of the 58th Annual Meeting of the Association
  for Computational Linguistics, {ACL} 2020, Online, July 5-10, 2020}, pages
  5821--5831.

\bibitem[{Song et~al.(2019)Song, Zhang, Cui, Wang, and
  Liu}]{song_exploiting_2019}
Haoyu Song, Wei-Nan Zhang, Yiming Cui, Dong Wang, and Ting Liu. 2019.
\newblock \href {https://doi.org/10.24963/ijcai.2019/721} {Exploiting {{Persona
  Information}} for {{Diverse Generation}} of {{Conversational Responses}}}.
\newblock In \emph{Proceedings of the {{Twenty-Eighth International Joint
  Conference}} on {{Artificial Intelligence}}}, pages 5190--5196, {Macao,
  China}. {International Joint Conferences on Artificial Intelligence
  Organization}.

\bibitem[{Song et~al.(2020{\natexlab{b}})Song, Zhang, Hu, and
  Liu}]{song_generating_2020}
Haoyu Song, Wei-Nan Zhang, Jingwen Hu, and Ting Liu. 2020{\natexlab{b}}.
\newblock \href {https://doi.org/10.1609/aaai.v34i05.6417} {Generating
  {{Persona Consistent Dialogues}} by {{Exploiting Natural Language
  Inference}}}.
\newblock \emph{Proceedings of the AAAI Conference on Artificial Intelligence},
  34(05):8878--8885.

\bibitem[{Vaswani et~al.(2017)Vaswani, Shazeer, Parmar, Uszkoreit, Jones,
  Gomez, Kaiser, and Polosukhin}]{NIPS2017_3f5ee243}
Ashish Vaswani, Noam Shazeer, Niki Parmar, Jakob Uszkoreit, Llion Jones,
  Aidan~N Gomez, \L~ukasz Kaiser, and Illia Polosukhin. 2017.
\newblock \href
  {https://proceedings.neurips.cc/paper/2017/file/3f5ee243547dee91fbd053c1c4a845aa-Paper.pdf}
  {Attention is all you need}.
\newblock In \emph{Advances in Neural Information Processing Systems},
  volume~30. Curran Associates, Inc.

\bibitem[{Vedantam et~al.(2015)Vedantam, Zitnick, and
  Parikh}]{vedantam2015cider}
R.~Vedantam, C.~Zitnick, and D.~Parikh. 2015.
\newblock Cider: Consensus-based image description evaluation.
\newblock In \emph{CVPR}, pages 4566--4575, Los Alamitos, CA, USA. IEEE
  Computer Society.

\bibitem[{Wolf et~al.(2020)Wolf, Debut, Sanh, Chaumond, Delangue, Moi, Cistac,
  Rault, Louf, Funtowicz, Davison, Shleifer, von Platen, Ma, Jernite, Plu, Xu,
  Scao, Gugger, Drame, Lhoest, and Rush}]{wolf-etal-2020-transformers}
Thomas Wolf, Lysandre Debut, Victor Sanh, Julien Chaumond, Clement Delangue,
  Anthony Moi, Pierric Cistac, Tim Rault, Rémi Louf, Morgan Funtowicz, Joe
  Davison, Sam Shleifer, Patrick von Platen, Clara Ma, Yacine Jernite, Julien
  Plu, Canwen Xu, Teven~Le Scao, Sylvain Gugger, Mariama Drame, Quentin Lhoest,
  and Alexander~M. Rush. 2020.
\newblock \href {https://www.aclweb.org/anthology/2020.emnlp-demos.6}
  {Transformers: State-of-the-art natural language processing}.
\newblock In \emph{Proceedings of the 2020 Conference on Empirical Methods in
  Natural Language Processing: System Demonstrations}, pages 38--45, Online.
  Association for Computational Linguistics.

\bibitem[{Wolf et~al.(2019)Wolf, Sanh, Chaumond, and
  Delangue}]{wolf_transfertransfo_2019}
Thomas Wolf, Victor Sanh, Julien Chaumond, and Clement Delangue. 2019.
\newblock {{TransferTransfo}}: {{A Transfer Learning Approach}} for {{Neural
  Network Based Conversational Agents}}.
\newblock \emph{arXiv:1901.08149 [cs]}.

\bibitem[{Wu et~al.(2020{\natexlab{a}})Wu, Li, Wang, Chen, Wong, Feng, Huang,
  and Wang}]{acl20-persona-variational}
Bowen Wu, Mengyuan Li, Zongsheng Wang, Yifu Chen, Derek~F. Wong, Qihang Feng,
  Junhong Huang, and Baoxun Wang. 2020{\natexlab{a}}.
\newblock \href {https://doi.org/10.18653/v1/2020.acl-main.7} {Guiding
  variational response generator to exploit persona}.
\newblock In \emph{Proceedings of the 58th Annual Meeting of the Association
  for Computational Linguistics, {ACL} 2020, Online, July 5-10, 2020}, pages
  53--65.

\bibitem[{Wu et~al.(2021)Wu, Zheng, Mao, and Huang}]{wu-etal-2021-transferable}
Chen~Henry Wu, Yinhe Zheng, Xiaoxi Mao, and Minlie Huang. 2021.
\newblock \href {https://doi.org/10.18653/v1/2021.emnlp-main.183} {Transferable
  persona-grounded dialogues via grounded minimal edits}.
\newblock In \emph{Proceedings of the 2021 Conference on Empirical Methods in
  Natural Language Processing}, pages 2368--2382, Online and Punta Cana,
  Dominican Republic. Association for Computational Linguistics.

\bibitem[{Wu et~al.(2020{\natexlab{b}})Wu, Hoi, Socher, and
  Xiong}]{wu-etal-2020-tod}
Chien-Sheng Wu, Steven~C.H. Hoi, Richard Socher, and Caiming Xiong.
  2020{\natexlab{b}}.
\newblock \href {https://doi.org/10.18653/v1/2020.emnlp-main.66} {{TOD}-{BERT}:
  Pre-trained natural language understanding for task-oriented dialogue}.
\newblock In \emph{Proceedings of the 2020 Conference on Empirical Methods in
  Natural Language Processing (EMNLP)}, pages 917--929, Online. Association for
  Computational Linguistics.

\bibitem[{Zhang et~al.(2018{\natexlab{a}})Zhang, Dinan, Urbanek, Szlam, Kiela,
  and Weston}]{zhang_personalizing_2018}
Saizheng Zhang, Emily Dinan, Jack Urbanek, Arthur Szlam, Douwe Kiela, and Jason
  Weston. 2018{\natexlab{a}}.
\newblock \href {https://doi.org/10.18653/v1/P18-1205} {Personalizing
  {{Dialogue Agents}}: {{I}} have a dog, do you have pets too?}
\newblock In \emph{Proceedings of the 56th {{Annual Meeting}} of the
  {{Association}} for {{Computational Linguistics}} ({{Volume}} 1: {{Long
  Papers}})}, pages 2204--2213, {Melbourne, Australia}. {Association for
  Computational Linguistics}.

\bibitem[{Zhang et~al.(2018{\natexlab{b}})Zhang, Galley, Gao, Gan, Li,
  Brockett, and Dolan}]{zhang_entropy}
Yizhe Zhang, Michel Galley, Jianfeng Gao, Zhe Gan, Xiujun Li, Chris Brockett,
  and Bill Dolan. 2018{\natexlab{b}}.
\newblock \href
  {https://proceedings.neurips.cc/paper/2018/hash/23ce1851341ec1fa9e0c259de10bf87c-Abstract.html}
  {Generating informative and diverse conversational responses via adversarial
  information maximization}.
\newblock In \emph{NIPS 2018}, pages 1810--1820.

\bibitem[{Zhao et~al.(2021)Zhao, Wallace, Feng, Klein, and
  Singh}]{zhao2021calibrate}
Zihao Zhao, Eric Wallace, Shi Feng, Dan Klein, and Sameer Singh. 2021.
\newblock Calibrate before use: Improving few-shot performance of language
  models.
\newblock In \emph{International Conference on Machine Learning}, pages
  12697--12706. PMLR.

\end{thebibliography}
\bibliographystyle{acl_natbib}

\appendix

\end{document}